\documentclass[runningheads]{llncs}
\usepackage[T1]{fontenc}
\usepackage{graphicx}

\usepackage{graphics} 
\usepackage{epsfig} 
\usepackage{mathptmx} 
\usepackage{times} 
\usepackage{amsmath} 
\usepackage{amssymb}  
\usepackage{subcaption}
\usepackage[normalem]{ulem}
\usepackage{todonotes}
\usepackage{hyperref}
\usepackage{setspace}
\usepackage[noend]{algpseudocode}
\usepackage[linesnumbered, ruled, vlined]{algorithm2e}

\begin{document}
\title{Biologically Inspired Neural Path Finding}
%
%
\author{Hang Li$^{\star}$\inst{1,3} \and
Qadeer Khan\thanks{ These authors contributed equally.}\inst{2} \and
Volker Tresp\inst{1,3}\and
Daniel Cremers\inst{2}
}

\authorrunning{Li et al.}
%
\institute{Ludwig Maximilian University of Munich \and 
\maketitle              
\begin{abstract}
The human brain can be considered to be a graphical structure comprising of tens of billions of biological neurons connected by synapses. It has the remarkable ability to automatically re-route  information flow through alternate paths in case some neurons are damaged. Moreover, the brain is capable of retaining information and applying it to similar but completely unseen scenarios. In this paper, we take inspiration from these attributes of the brain, to develop a computational framework to find the optimal low cost path between a source node and a destination node in a generalized graph. We show that our framework is capable of handling unseen graphs at test time. Moreover, it can find alternate optimal paths, when nodes are arbitrarily added or removed during inference, while maintaining a fixed prediction time.  Code is available here: \url{https://github.com/hangligit/pathfinding}

\end{abstract}
\section{Introduction}\label{sec:introduction}
We are inundated with graphical structures of various forms in this contemporary era of digitization. This includes, for e.g., social networks \cite{leskovec_rajaraman_ullman_2014} wherein the nodes represent the individuals and the edges characterize the social connections between the individuals. Another popular form of network includes recommender systems \cite{ying2018graph} that can be represented as bipartite graphs. The users/products represent the nodes, while the edges depict the rating of likes/dislikes of a user for a certain product. Other graphs include citation networks \cite{sen2008collective}, molecular structures used in drug discovery \cite{gilmer2017neural} etc.

Although concurrent implementation of these graphical structures are computationally powerful in number-crunching, they lack the cognitive understanding to draw meaningful conclusions that can readily be interpreted.  On the other hand, one of the most sophisticated and yet least understood graphical networks is the human brain \cite{peer2021structuring}. Rather than consisting of computational nodes, it is comprised of tens of billions of biological neurons both sending and receiving information to the neighbouring neurons through the connecting synapses \cite{azevedo2009equal,bastos2012canonical}. One amazing attribute of the human brain is its ability to learn to automatically adapt and efficiently reroute information through alternate neural paths, in case of certain damaged neurons \cite{zelikowsky2013prefrontal}. Another important attribute is the capability to interpret distinguishing patterns in data and retain this information to be applied in similar circumstances in the future \cite{lake2017building}. For e.g., a child touching a hot cup of coffee once or twice would feel a sensation of pain. The child's brain will retain this experience to avoid touching hot cups in the future even if the cups are of different size/colour/shape etc. However, unlike computers whose computation power has been exponentially rising over the past 4 decades, the capacity of the human brain to process information is limited by biological constraints.  Therefore, is it possible to combine the benign attributes of the human brain with the processing power of computational resources? In this work, we explore this possibility in the context of path optimization.  

The ability to navigate through a network from a source to a destination node while optimizing for the lowest cost is an important problem. It has a tremendous number of diverse applications, for e.g., the ubiquitous vehicle/robot navigation. The cost could involve minimizing either the distance travelled, time taken, or even the traffic congestion encountered. Other less frequent, but critical, use cases are search and rescue operations involving unmanned aerial vehicles. Here, minimizing the battery usage and the data transmission are important factors to be optimized for. Traditionally, these problems can either be solved heuristically using approaches such as A-star or by deploying \emph{"shortest path"} algorithms such as Depth First Search (DFS), Breadth-First Search (BFS), Djikstra, etc. These approaches tend to start with the source node and progressively traverse the graph through the neighbouring nodes, then neighbours of the neighbours until the destination node is found.  Although accurate, their computational complexity rises with the number of hops between the source and destination nodes. On the other hand, given a visual map drawn to scale, humans are fairly good at quickly determining the approximate optimal path\cite{bongiorno2021vector}, irrespective of the number of hops between the nodes. Is it also possible to additionally emulate this one-shot prediction capability in a computational setting? In this regard, we propose using Graphical Neural Networks (GNNs) to find the path with the lowest cost. Our framework has the following attributes:

\begin{enumerate}
    \item If a node(s) or edge(s) is arbitrarily removed from the graph structure, the optimal path is automatically rerouted through the remaining nodes/edges to find the next best solution.
    \item The framework can generalize to find the optimal path even on unseen graphs.
    \item The time taken to find the lowest cost path between the source and destination node remains constant irrespective of the number of hops between them.

\end{enumerate}
\section{Related Work}\label{sec:related_work}
\noindent{\textbf{Artificial Neural Networks, ANNs:}}\\ Over the last decade, the advent of data-driven, learning-based methods for training artificial neural networks has  made  tremendous  strides  in  achieving unprecedented levels of performance on various tasks such as natural language processing \cite{radford2019language}, computer vision \cite{NIPS2012_c399862d}, medical diagnosis \cite{golkov}, etc. Such networks have the capability to extract meaningful information from the training data and extrapolate this to completely unseen test data. In fact, they have achieved on par human performance \cite{CIRESAN2012333} on tasks such as classification, disease diagnosis, etc. We also deploy ANNs to achieve generalization on unseen data (graphs).\\

\noindent{\textbf{Graph Representation:}}\\ Among the various ANN paradigms, feed-forward architectures such as Convolutional Neural Networks (CNNs),  Multi-layer perceptrons etc. are ubiquitous. However, on tasks such as path optimization, there is ambiguity on how the graph structure should be represented when input through such architectures. One approach is to input the graph as an adjacency matrix. \cite{8363534} use the connectome \cite{201689} as input to a CNN to predict autism in patients. The connectome is an adjacency matrix, encoding brain connectivity as graphs between certain pre-selected regions in the brain. \cite{KAWAHARA20171038} used a pre-determined number of regions in the connectome to train BrainNetCNN for predicting neurodevelopment. However, one major limitation of using adjacency matrices is that the number of nodes forming the input to the network cannot change at inference time. Otherwise, the network needs to be trained from scratch if the number of nodes is increased. Our framework does not suffer from this limitation and the number of nodes forming the input to the network can be changed without retraining the network. In fact, we show in the experiments that our method is capable of automatically rerouting to an alternate path, if some nodes/edges are removed at test time. \\

\noindent{\textbf{Reinforcement Learning (RL):}}\\ RL is an alternate strategy for determining the optimal path in a map/graph. Taking inspiration from human psychology, a reward function is defined \cite{NIV2009139,PALMINTERI2013131}. The agent explores the environment in a hit-and-trial manner incurring rewards along the way \cite{sutton2018reinforcement}. If the training parameters are carefully chosen, the agent converges to an optimal policy. \cite{PANOV2018347} demonstrated path planning on small unseen maps. Instead of the adjacency matrix, they directly used the map represented as a grid. While this can handle maps with arbitrary structure and nodes, it is constrained to only planar graphs. In contrast, our framework can additionally handle graphs with non-planar structure.\\

\noindent{\textbf{Shortest Path Algorithms:}}\\ Shortest path algorithms such as DFS, BFS, Djikstra \& their modifications have been used in a wide array of applications. These range from fast IP rerouting \cite{8321810}, to marine navigation \cite{9164597}, software defined networking \cite{6996609}, maze solving \cite{8226690} to even optimal planning of sales persons \cite{7522330}. The limitation of these methods is that the time to find the optimal path is not constant. Rather it depends on the number of nodes \& edges in the graph. For dense graphs, the computational complexity can be of polynomial order of the number of nodes. On the other hand, the computational complexity of our method is constant. Irrespective of the number of hops between the source and  destination nodes, our framework takes the same time to find the optimal path.  
\section{Framework}\label{sec:method}
The task tackled here is to find the lowest cost path between a source and a destination node for an undirected graphical structure, $G$. The nodes are connected through edges having arbitrary costs. The edge $e\textsubscript{i,j}\in E$ connects node $i$ ($v\textsubscript{i} \in V$) to node $j$ ($v\textsubscript{j} \in V$). Here $V$ and $E$ are the sets containing all nodes and edges respectively. 

In our framework, we use ANNs, as they are capable of learning patterns in the data and robustly applying them to unseen data. However, in the context of optimal path finding on graphs, we would like our framework to possess two additional properties:
\begin{enumerate}
    \item It should be invariant to the permutations/ordering of the nodes.
    \item Addition/removal of nodes should not render the training of the ANN useless. Rather it should be capable of being trained \& tested on any number of nodes.
    
\end{enumerate}

Traditional feedforward networks such as MLPs do not possess these two important attributes.  They are susceptible to node ordering and cannot easily handle addition of new nodes. Figure \ref{fig:feedforward_problem} depicts the implications on the adjacency matrix, when the node order is changed and when a new node is added. In the node permutation case, the adjacency matrix is completely different, despite the graphs being isomorphic. In the scenario of node addition, the adjacency matrix is extended. To accommodate this extension, the architecture of the MLP would also need to be changed; thereby requiring re-training. 

\begin{figure}[h]
    \centering
    \includegraphics[width=1.0\textwidth]{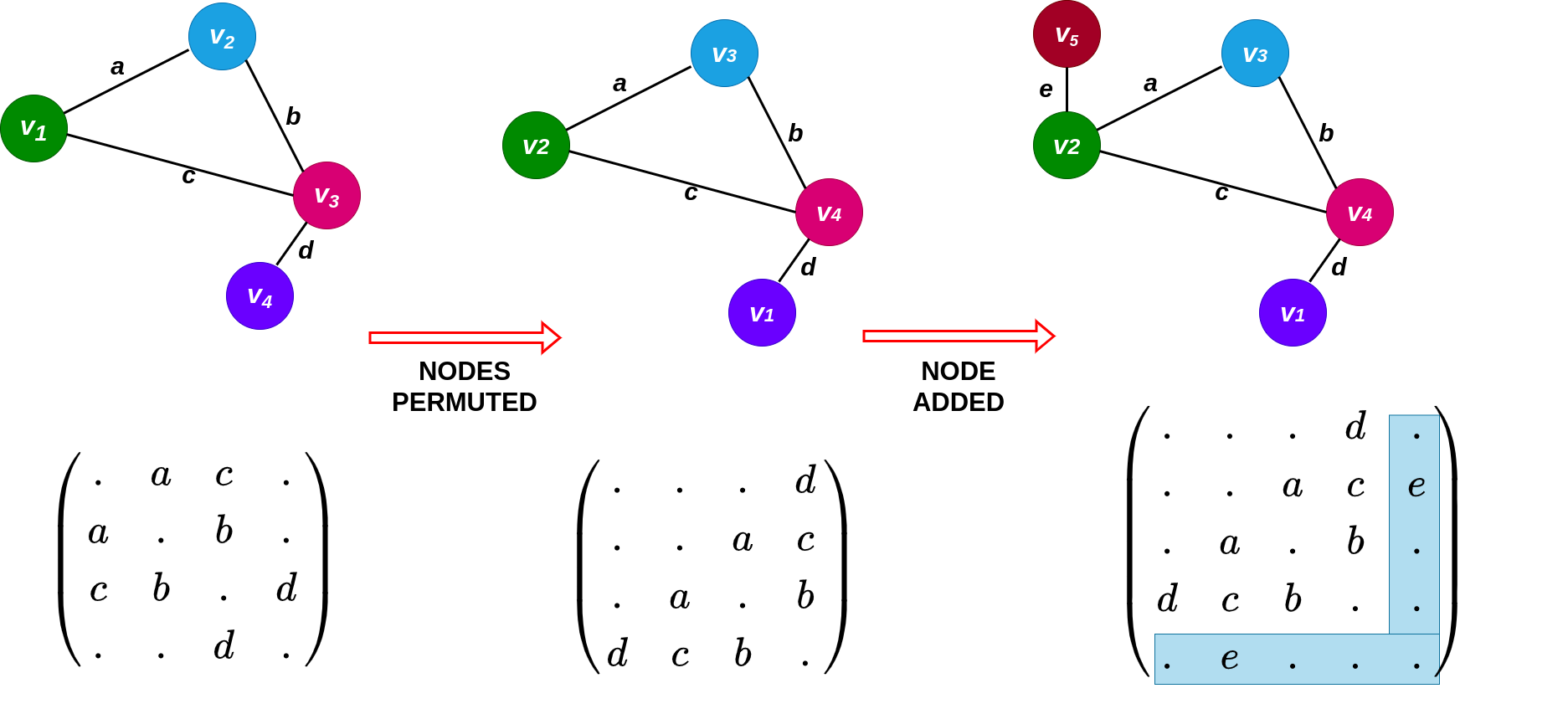}
    \caption{Shows the implications of permuting the node order and adding an additional node on the adjacency matrix of a graph.}
    \label{fig:feedforward_problem}
\end{figure}
To circumvent these issues, we also use MLPs, but in the paradigm of a Graphical Neural Network. Let the  input features for a node $i$ be represented by ${z}^{(0)}_{i}$. These input features for each node are passed through a series of graphical layers to produce latent embeddings (${z}^{(l)}_{i}$, for node $i$ at layer $l$). Figure \ref{fig:gnn_basics} depicts a high level overview of the constituents of layer $l$. For simplicity, we demonstrate the information flow for the first graph shown in Figure \ref{fig:feedforward_problem}. Note that the layer $l$ takes the embeddings (${z}^{(l-1)}$) for each node from the previous layer $l-1$ as input to produce corresponding embeddings ${z}^{(l)}$ as output. The inputs are first passed through a neural network ($M$) comprised of fully connected layer(s) followed by a non-linearity. An important characteristic of this neural network is that the weights are shared between all nodes. Hence, nodes can conveniently be added or removed from a graph without re-training the network. This is because the added node can simply use the same weights as those of the other nodes. The output of the neural network is then passed through an aggregation function. For each node, the input to the aggregation function depends on which are its neighbours. For e.g., the neighbours of $v_1$ are $v_2$ and $v_3$. Hence, information from $M$(${z}^{(l-1)}_{2}$)  and $M$(${z}^{(l-1)}_{3}$) is aggregated to produce ${z}^{(l)}_{1}$. The aggregation function is chosen so that its output is invariant to the order of the input. Hence, even if the ordering of nodes in a graph is changed, the  output remains the same. Some examples of order invariant aggregation functions include summation, mean, taking the maximum/minimum etc. for each scalar value of the input vectors.
Note that in the first layer, each node would incorporate information from its immediate neighbours, the next layer will implicitly draw information from the neighbours of its neighbours. The deeper we go into the network, each node will retrieve information from nodes farther away from it in the graph. 

The output embedding of the last graph layer is then passed through to a classifier which predicts whether or not the corresponding node falls within the optimal path. Next, we describe the mathematical details of the graph layers, the input node features, the loss function to train the weights and how the edge weights are incorporated into the graph based upon a modification of \cite{velickovic2017graph}. 
\begin{figure}[h]
    \centering
    \includegraphics[width=1.0\textwidth]{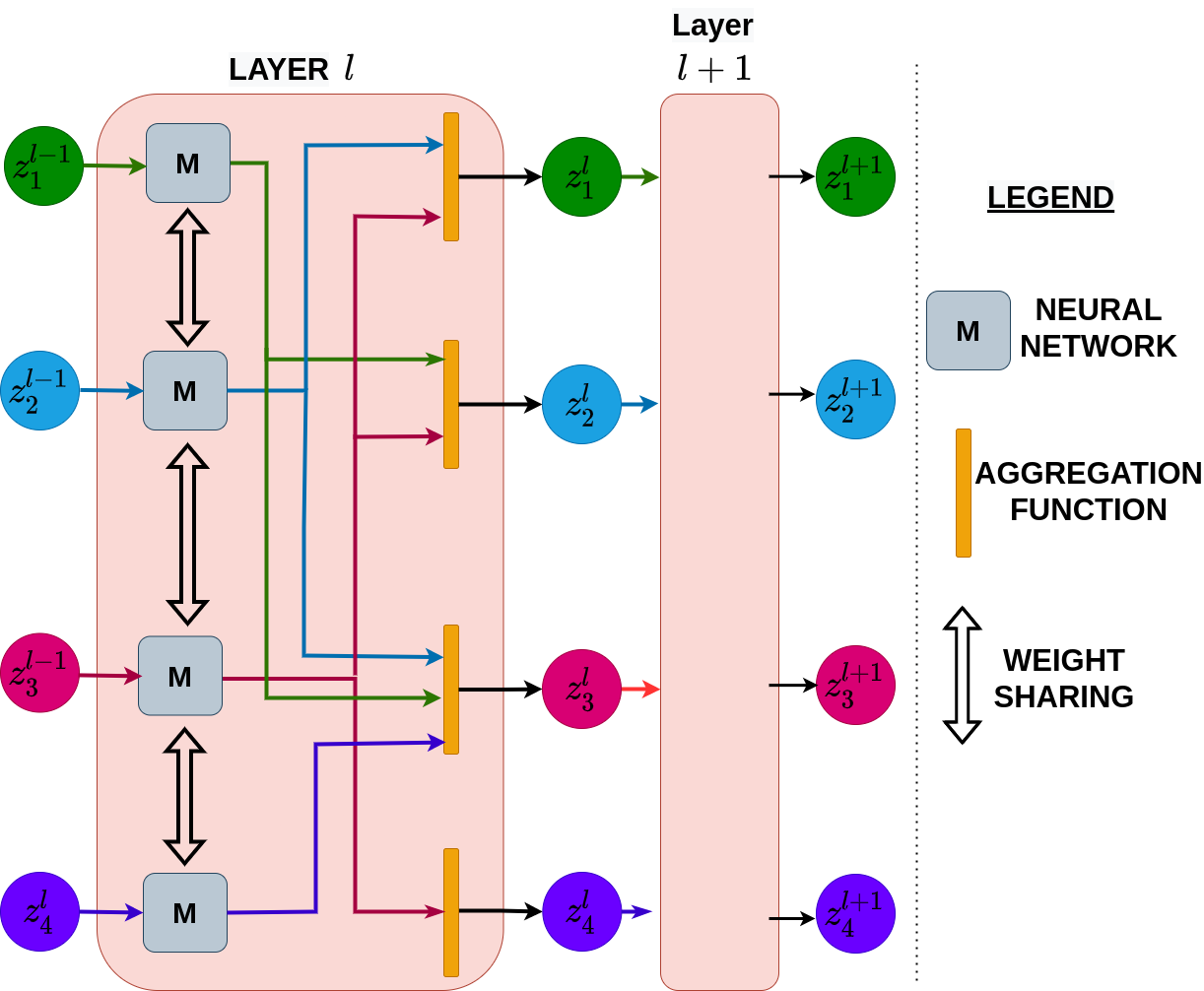}
    \caption{Gives the high-level overview of graphical layer $l$, for the first graph shown in Figure \ref{fig:feedforward_problem}. Node embeddings (${z}^{(l-1)}$) from the previous $l-1$ layer are taken as input to produce node embeddings ${z}^{(l)}$ which can then be fed to the next $l+1$ layer to produce embedding ${z}^{(l+1)}$ and so on. }
    \label{fig:gnn_basics}
\end{figure}
\newline
The embedding $z^{(l)}_i$ $\in \mathbb{R}^{d_{l}}$ of a node $i$ at the $l$-th layer is updated as:

\begin{equation}
    z^{(l)}_{i} = \mathrm{D}(\mathrm{\sigma}({\hat{z}}^{(l)}_{i})).
\end{equation}
Here, $\sigma$ is the non-linearity (we use LeakyReLU). $D$ is the dropout layer. Note that $\hat{z}$ is obtained from the input node features of the previous layer, i.e., ${z}^{(l-1)}_i(\in \mathbb{R}^{d_{l-1}}$)  in the following manner:
\begin{equation}
  {\hat{z}}^{(l)}_{i} = \sum_{j \in \mathcal{N}(i) \cup \{i\}} \alpha_{ij} \left( \mathbf{W}^{(l)} {z}_j^{(l-1)} \right)
\end{equation}
where ${W}^{(l)} \in \mathbb{R}^{d_{(l)} \times  d_{(l-1)}} $ are the trainable parameters of the neural network. Meanwhile,  $\mathcal{N}(i) \in V$ is the set of neighboring nodes of $i$ across which aggregation is done through summation. The scalar weights $\alpha_{ij}$ incorporate the edge cost between nodes $i$ and $j$ by way of this equation:
\begin{equation}
    \alpha_{ij} = \frac{\exp(\mathbf{a}^{\top}\mathrm{\sigma}([\mathbf{W}^{(l)}{z}_i||\mathbf{W}^{(l)}{z}_j||\mathbf{W^{e}}e_{ij}]))}{\sum_{j^{'}\in \mathcal{N}(i) \cup \{i\}} \exp(\mathbf{a}^{\top}\mathrm{\sigma}(\mathbf{W}^{(l)}{z}_i||\mathbf{W}^{(l)}{z}_{j^{'}}||\mathbf{W^{e}}e_{ij^{'}})).}
\end{equation}
The edge weights $e_{ij}$ are first mapped to a $d_{(l)}$ dimensional hidden feature representation $h_{ij}=\mathbf{W^{e}}e_{ij}$ where $\mathbf{W^{e}}\in \mathbb{R}^{d_{(l)}}$ . The \textbf{||} symbol represents a concatenation of vectors. $\mathbf{a} \in \mathbb{R}^{3 d_{(l)}}$ and $\mathbf{W^{e}}$ are trainable parameters.

Note ${z}^{(0)}_i\in \mathbb{R}^{3}$ are the one-hot encoded input features representing whether a node in the graph is either the source, the destination or otherwise.

The output from the final graph layer $L$ is the node embedding ${z}^{(L)}_i$. In the final layer, we also determine the edge embedding (${u}_{ij}$) for each edge . It is the element-wise sum of embeddings of nodes that it connects to, i.e., $u_{ij}={z}^{(L)}_i+{z}^{(L)}_j$

These edge embeddings and final layer node embeddings are passed through their respective MLP classification layers. It predicts the probabilities of them being in the optimal path. This probability for the node $i$ and edge $ij$ are respectively given by the following equations:
\begin{equation}
    \hat{p_i} = \mathrm{\sigma_{2}}(\mathbf{W}^{n}_2(\sigma(\mathbf{W}^{n}_1{z}_i^{(L)}+\mathbf{b}^{n}_1)+\mathbf{b}^{n}_2))
\end{equation}
\begin{equation}
    \hat{p_{ij}} = \mathrm{\sigma_{2}}(\mathbf{W}^{e}_2(\sigma(\mathbf{W}^{e}_1u_{ij}+\mathbf{b}^{e}_1)+\mathbf{b}^{e}_2)).
\end{equation}

\noindent{$\mathbf{W}^{n}_1, \mathbf{W}^{e}_1 (\in \mathbb{R}^{m \times {d_{(L)}}})$,  $\mathbf{W}^{n}_2, \mathbf{W}^{e}_2 (\in \mathbb{R}^{1 \times m})$, $\mathbf{b}^{e}_1$, $\mathbf{b}^{n}_1 (\in \mathbb{R}^m)$,  $\mathbf{b}^{e}_2$, $\mathbf{b}^{n}_2 (\in \mathbb{R})$ are also the trainiable parameters of the model. $m$ is a hyper-parameter. $\sigma_{2}$ is the sigmoid non-linearity. }

\noindent{One} may ask why we need separate weights for the node \& edge classifications. Or even why is the edge classification needed at all? This is because two nodes being in the optimal path does not imply that the edge connecting them will necessarily be in the optimal path.  Figure \ref{fig:node_removal} describes a simple example demonstrating the importance of additionally predicting the probability of the edge being in the optimal path. Although, nodes 3 \& 4 are in the optimal path, but the edge connecting them (cost=8) is not. We also empirically found that predicting the probability for both edges \& nodes is better than predicting for only the nodes or only the edges.

The loss function is the binary cross entropy between the predicted probability and the ground truth over all training samples
$$\min_{\mathbf{W},\mathbf{b},\mathbf{a}} \mathbb{E}_{G,y \sim \mathrm{Data}} -[ y \log \hat{p} + (1-y) \log(1-\hat{p}) ].$$ 
The ground truth can be obtained directly while constructing a random graph structure of larger cost around the optimal path. Or using shortest-path algorithms on known graphs.

\section{Experiments}\label{sec:experiments}

Our model is trained with a  learning rate of 1e-4 with the Adam optimizer used for updating the weights. The data comprises of a total 10000 different arbitrarily created graphical structures with up to 30 nodes. The training, validation and test split is [0.7:0.15:0.15]. The test set is comprised of graphs with an arbitrary number of nodes. This serves to analyze if the model can handle different number of nodes in the inference graph. The test set also contains graph samples wherein some nodes/edges have been arbitrarily removed. This is to see if the model can determine alternate paths in case of such removal. Also, note that each of the 10000 structures is comprised of 10 different edge weight combinations for a total of 100000 samples. The ground truth labels for the loss function are obtained using \cite{dijkstra1959note}.  \\

\noindent\textbf{{Evaluation Metric:}}\\The metric we use for quantitative evaluation is the \emph{Path Accuracy}. It is the ratio of the number of graph samples in the unseen test set, wherein the class of every node/edge in the graph is correctly predicted. Hence, not only every node/edge in the optimal path must be classified as such but the nodes/edges not in the optimal path should also be correctly classified as not belonging to the optimal trajectory. Table \ref{tab:results} reports the \emph{Path Accuracy} metric on both the training and unseen test set for our method and its variations. This is further elaborated in the next Section \ref{sec:discussion}. 
\begin{center}

\begin{tabular}{||c c c c c c||}
 \hline
  & Ours & \shortstack{Fixed \\ Structure} & \shortstack{Fixed \\ Nodes} & \shortstack{Nodes \\ Only}  & \shortstack{Edge \\ Only} \\ [0.5ex] 
 \hline\hline
 \shortstack{Training \\ Data} & 98.01 & 99.26 & 99.41  & 97.75 & 97.43 \\ 
 \hline
 \shortstack{Test \\ Data (Unseen)}  & \textbf{98.02} & 68.41 & 72.18  & 97.62 & 97.41\\
 \hline
\end{tabular}
\captionof{table}{Reports the \emph{Path Accuracy} metric for our method and its variations for both the training and test set. (Higher is better) }
\label{tab:results}
\end{center}

\section{Discussion}\label{sec:discussion}
We now give a more detailed analysis of the results from our experimental evaluation.

\noindent{\textbf{Unseen test data:}} \\
From Table \ref{tab:results} it can be seen that our model is capable of maintaining good \emph{Path Accuracy} performance even on unseen test data. Note that the test data comprises of graph samples with an arbitrary number of nodes between 3 and 50. In addition, it also contains graph samples wherein the edges/nodes are arbitrarily removed. Our model is capable of robustly handling both scenarios. Figure \ref{fig:node_variability} shows a plot of the accuracy as the number of nodes is changed. Note that as the number of nodes in the graph is changed, the accuracy as depicted by the green curve remains fairly consistent. This is because our model is trained to handle such instance with variable number of nodes in the graph. 

\begin{figure}[!h]
    \centering
    \includegraphics[width=\textwidth]{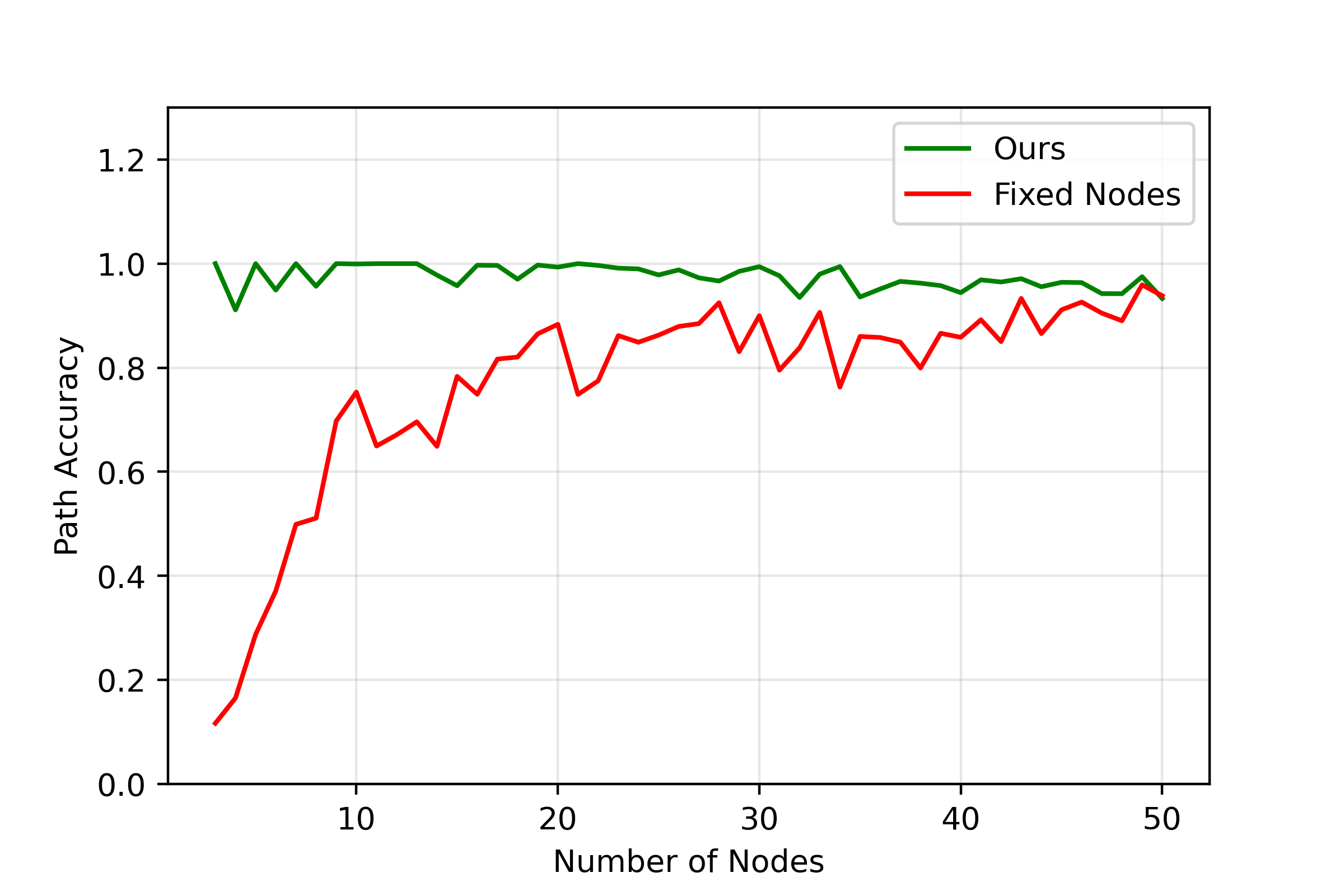}
  \caption{The left plot shows that, as the number of nodes is changed, the path accuracy metric of our model (in green) remains constant. It is interesting to note that even though the model was trained with up to a maximum of 30 nodes, it still maintains good performance beyond this number. The red curve on the plot shows the performance for a model trained with a fixed number of nodes. Its performance deteriorates when evaluated on a smaller number of nodes. }
    \label{fig:node_variability}
\end{figure}

Meanwhile, Figure \ref{fig:node_removal} demonstrates a simple example of finding an optimal alternate path in the case of a removed edge. Note that initially in the original graph, the optimal path between nodes 2 and 4 is the direct edge connecting the two nodes having a cost of 1. However, when this removed is removed the alternate lowest cost path between the nodes is 2-0-1-3-5-4. This alternate has a cost of 7 which is the lowest in the graph after the edge removal.

\begin{figure}[!h]
    \centering
    \includegraphics[width=\textwidth]{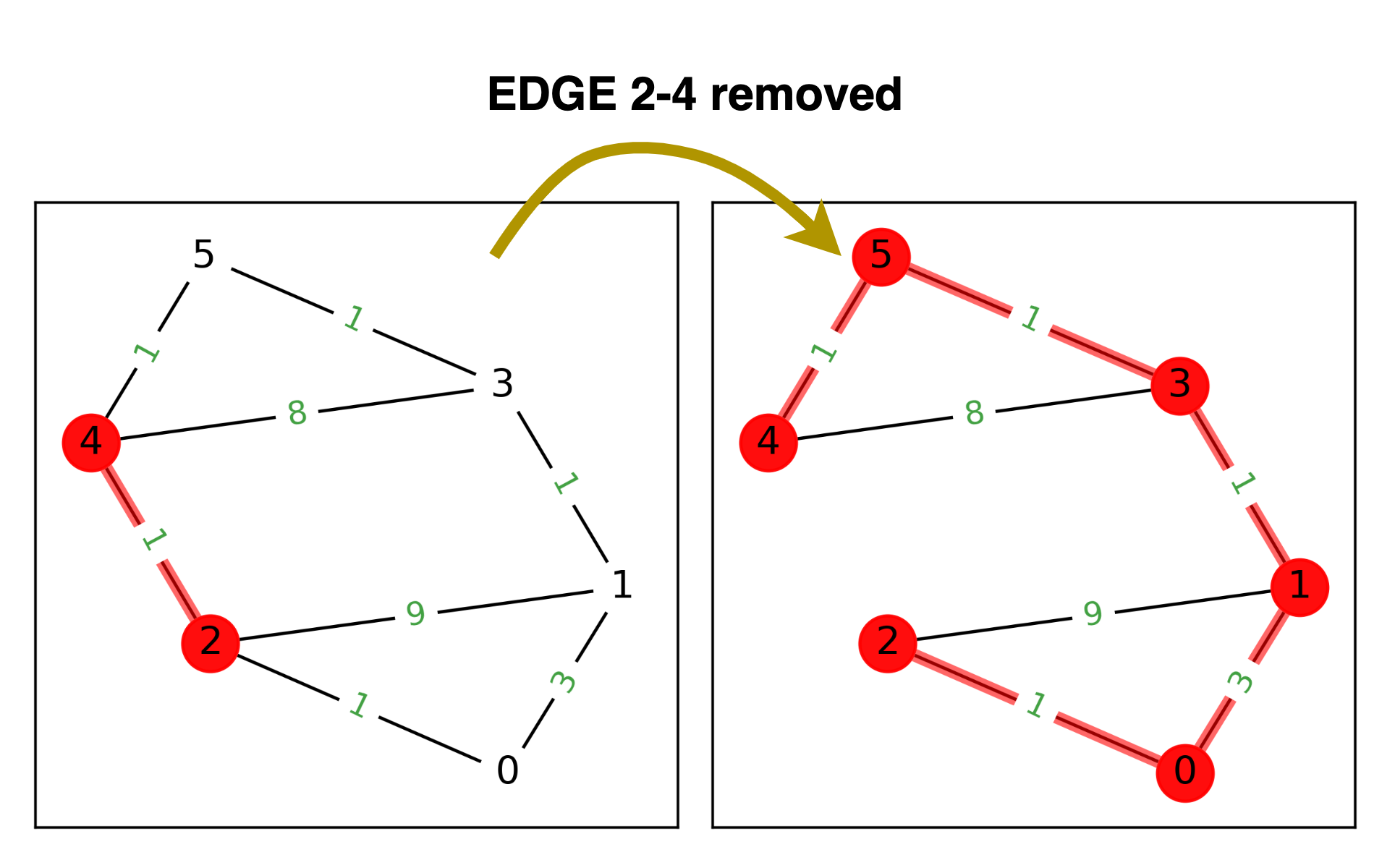}
  \caption{The Figure shows that if the shortest path between nodes 2-4 is removed, the network is capable of automatically finding the alternate shortest path through nodes 2-0-1-3-5-4. The cost via this alternate path is the lowest in the modified graph resulting from the edge removal }
    \label{fig:node_removal}
\end{figure}

\noindent{\textbf{Fixed structure, fixed number of nodes:}}\\ Note that in our framework not only the number of nodes can change, but also the structure of the graph constituted by the same nodes can vary.  To demonstrate this, we train another model wherein both the structure and the number of nodes (30) are kept constant. Only the edge weights are changed. Consider the 2$^{nd}$ column of Table \ref{tab:results} for the results. While the model performs well on the training samples containing graphs of the same structure \& nodes, its performance on the unseen test data drops dramatically. This is because when training the model, it is not accustomed to handling variable nodes \& structures of the test data.

\noindent{\textbf{Variable structure, fixed number of nodes:}} \\
To compensate for this, we train another model, with the same number of nodes, but for which the structure of the graph can change. Note that the performance of this model is good even when the structure is changed at test time. This is because the training incorporated changing structure into the model. However, because the number of nodes is fixed, its performance drastically starts to drops when additonal nodes are added/removed to the graph as can be seen in the red curve in Figure \ref{fig:node_variability}. In contrast our approach can not only accomdate addition of new nodes but also changing structure of the graph while maintaining a stable performance.

\noindent{\textbf{Comparision with BrainNetCNN approach:}}\\ Note that the BrainNetCNN approach uses adjacency matrices, which requires filter sizes in the CNN to be fixed. Hence, graphs with an arbitrary number of nodes cannot readily be handled. The original BrainNetCNN was meant for classifying neurodevelopment in patients. Therefore, we slightly modify the deeper layers to classify nodes/edges in the optimal path instead. To further elaborate this, Figure \ref{fig:fig_brainnetcnn} shows a 3x3 grid structure depicting the accuracy distribution for 3 different models evaluated on 3 different datasets. Each column in the grid represents a different dataset and is described as follows:
\begin{enumerate}
\item \textit{Variable Weights:} All graphs in this dataset have the same structure but have different edge weights
\item  \textit{Node Permutations:} All graphs in this dataset have the same structure but have the nodes permuted and also different edge weights.
\item  \textit{Variable Structures:} The graphs in this dataset can have not have different structure, different edge weights. However, the number of nodes still remain the same. 
\end{enumerate}
Each row represent a different model, training of which is described as follows:
\begin{enumerate}
\item \textit{Variable Weights:} This model was trained with graphs with on graphs having same structure but different edge weights. This model has good performance on the Variable Weights dataset but performs poorly when evaluated on the datasets wherein the nodes are permuted or where the structure is different. This is understandable because this model is not accustomed handling such graphs during training.
\item  \textit{Node Permutations:} This model was trained on graphs having the same structure but with the nodes permuted and also different edge weights. Here this model performs well on both the Variable Weights and Node Permutations dataset. However, the performance of this model drops when evaluated on graphs with variable structures.
\item  \textit{Variable Structures:} This model was trained on graphs having different structures and has a good, stable performance across all the 3 datasets.
\end{enumerate}

 \begin{figure}[!ht]
     \centering
     \includegraphics[width=\textwidth]{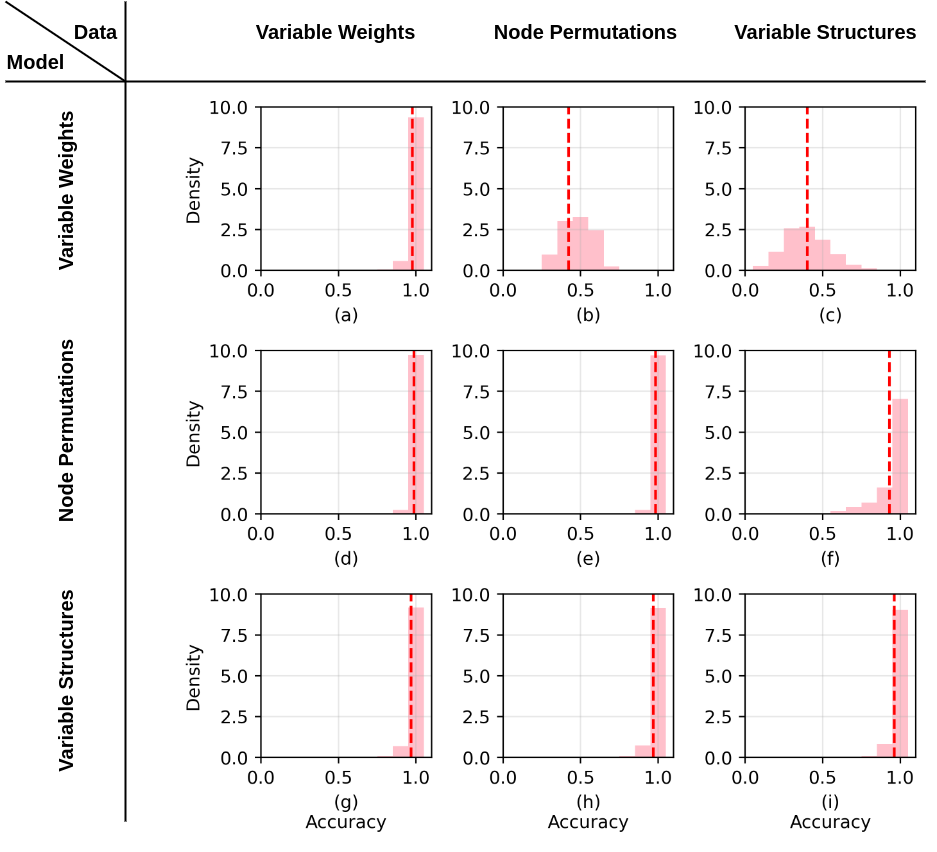}
     \caption{The first row shows the test performance of a BrainNetCNN trained with fixed structue and fixed number of nodes. Each coloumn from left to right denotes graphs with fixed structure, node permutations, and variable structures. The second row shows model trained with node permutations. The model trained with different structures (the third row) generalizes well across all the 3 datasets. It is important to mention that all datasets contained a fixed number of nodes and all models were also trained with a fixed number of nodes.}
     \label{fig:fig_brainnetcnn}
 \end{figure}

However, note that all these models could only be trained and evaluated on graphs with the same number of nodes. Our framework on the other hand does not suffer from this problem and can additionally be trained and evaluated on grpahs with a variable number of nodes as already depicted in Figure \ref{fig:node_removal}.

\noindent{\textbf{Relative Prediction time:}}\\ Figure \ref{fig:relative_time} reports the relative time to find the optimal path as the number of hops between the source and destination nodes is increased. It is normalized by the time taken to find the optimal path between nodes one hop away. As can be seen, this number remains stable for our approach. Hence, irrespective of the number of hops we have in the graph, the prediction time remains the same. This is because a forward pass of our model always involves the same number of graph conventional layers. Compare this with Djikstra's algorithm wherein the relative time increases with the number of hops.

\begin{figure}[!h]
    \centering
    \includegraphics[width=0.75\textwidth]{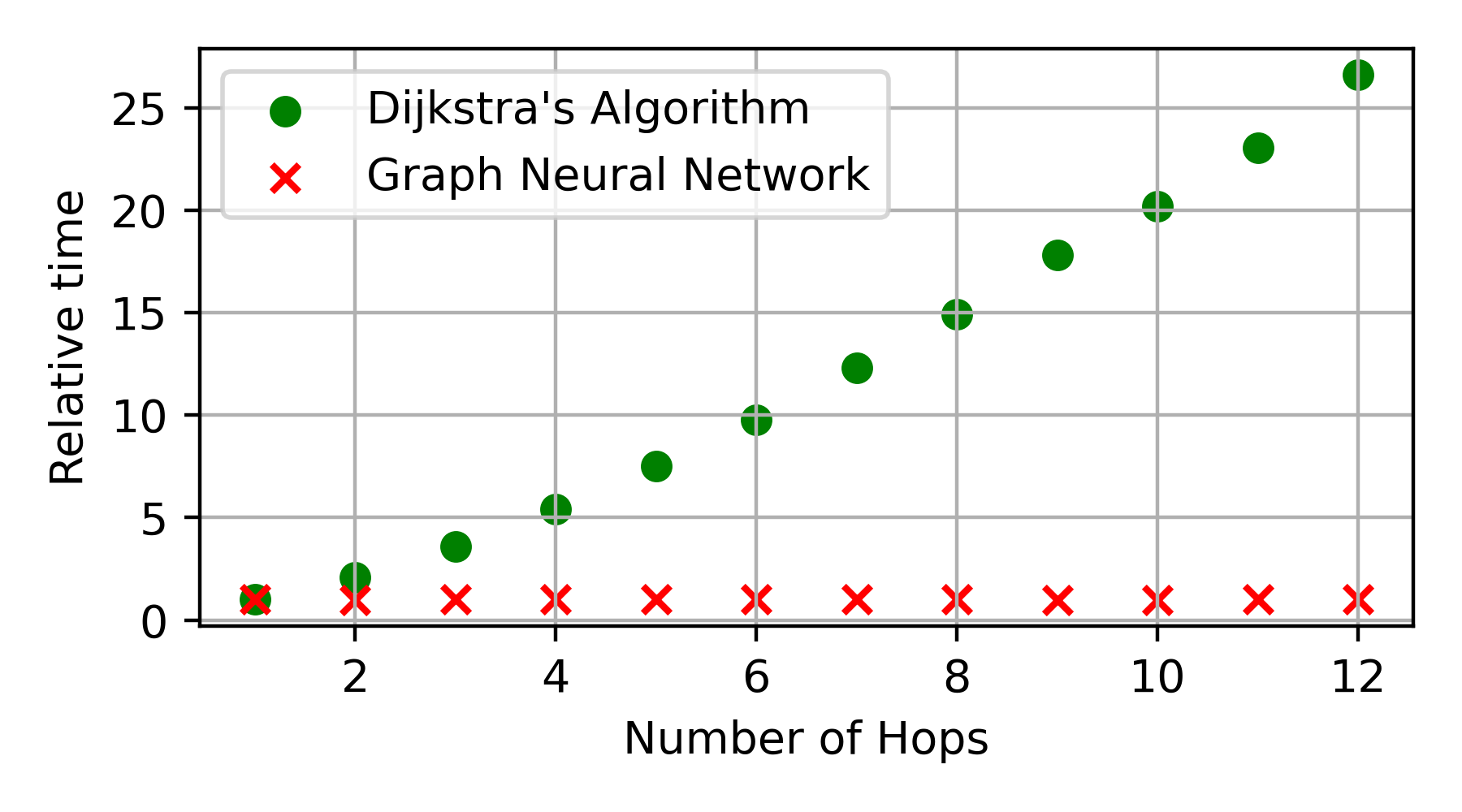}
  \caption{The plot shows the relative prediction time as a function of the number of hops between the source and destination nodes. }
    \label{fig:relative_time}
\end{figure}

\noindent{\textbf{Loss functions:}} \\
In addition to classifying the nodes in the optimal path, our loss function also incorporates classifying the edges connecting the nodes in the path. The last 2 columns in Table \ref{tab:results} shows the implications of training with the binary cross entropy function only for the nodes and only for the edges. As can be seen, the performance of our model, which combines both loss functions, is superior to the models trained with only the individual loss functions.

\noindent{\textbf{Evaluation on a real world dataset:}}\\ We also evaluated our approach of optimal path finding on maps of the real-world KITTI \cite{Geiger2012CVPR} dataset and found that it achieved a perfect score on the \emph{Path Accuracy} metric. One plausible explanation for this is that the maps of the road structure in KITTI are on a plane. Our model, in contrast, was trained with non-planar graphs which tend to be more challenging to handle and hence do not achieve a perfect score when evaluated on a test set comprising non-planar graphs.

\section{Conclusion}\label{sec:conclusion}
In this paper we demonstrated how our biologically inspired computational framework is capable of optimal path finding. It mimics the behaviour of the brain to find alternate shortest paths on unseen data even when nodes/edges are removed. This is unlike adjacency-matrix based conventional feedforward approaches which cannot be trained with varying numbers of nodes. As we developed our framework for generalized graph structures, it can be extended to various applications.

\section{Acknowledgement}
This work was supported by the Munich Center for Machine Learning and the BMBF-project MLWin.

This preprint has not undergone peer review (when applicable) or any post-submission improvements or corrections. The Version of Record of this contribution is published in Brain Informatics, and is available online at \sloppy \url{https://doi.org/10.1007/978-3-031-15037-1_27}.
\bibliographystyle{splncs04}

\bibliography{bibfile}

\end{document}